\documentclass[conference]{IEEEtran}
\IEEEoverridecommandlockouts
   
    \newcommand{\VT}[1]{\ensuremath{{V_{T#1}}}}

    \newcommand{\bfv}  	{\mathbf{v}}
    
    \newcommand{\bfx}  	{\mathbf{x}}
    \newcommand{\bfy}  	{\mathbf{y}}

    \newbox\sectsavebox
    \setbox\sectsavebox=\hbox{\boldmath\VT{xyz}}
    
\usepackage{cite}
\usepackage{amsmath,amssymb,amsfonts}
\usepackage{bm}
\usepackage{algorithmic}
\usepackage{graphicx}
\usepackage{adjustbox}
\usepackage{textcomp}
\usepackage{makecell}
\usepackage{xcolor}
\usepackage{comment}
\usepackage{mathtools}
\usepackage{subcaption}
\def\BibTeX{{\rm B\kern-.05em{\sc i\kern-.025em b}\kern-.08em
    T\kern-.1667em\lower.7ex\hbox{E}\kern-.125emX}}
\begin{document}

\title{Robust Semi-Supervised Classification using GANs with Self-Organizing Maps\\
}

\author{\IEEEauthorblockN{Ronald Fick\IEEEauthorrefmark{1}\IEEEauthorrefmark{5},
Paul Gader\IEEEauthorrefmark{2}\IEEEauthorrefmark{5}, and Alina Zare\IEEEauthorrefmark{3}\IEEEauthorrefmark{4}}
\IEEEauthorblockA{\IEEEauthorrefmark{5}Computer and Information Science and Engineering\\
\IEEEauthorrefmark{4}Electrical and Computer Engineering\\
\textit{University of Florida}, Gainesville, USA\\
Email: \IEEEauthorrefmark{1}rfick@ufl.edu,
\IEEEauthorrefmark{2}pgader@ufl.edu,
\IEEEauthorrefmark{3}azare@ufl.edu}}

\maketitle

\begin{abstract}
Generative adversarial networks (GANs) have shown tremendous promise in learning to generate data and effective at aiding semi-supervised classification. However, to this point, semi-supervised GAN methods make the assumption that the unlabeled data set contains only samples of the joint distribution of the classes of interest, referred to as inliers.  Consequently, when presented with a sample from other distributions, referred to as outliers, GANs perform poorly at determining that it is not qualified to make a decision on the sample. The problem of discriminating outliers from inliers while maintaining classification accuracy is referred to here as the DOIC problem.  In this work, we describe an architecture that combines self-organizing maps (SOMs)  with SS-GANS  with the goal of mitigating the DOIC problem and experimental results indicating that the architecture achieves the goal. Multiple experiments were conducted on hyperspectral image data sets. The SS-GANS performed slightly better than  supervised GANS on classification problems with and without the SOM. Incorporating the SOMs into the SS-GANs and the supervised GANS led to substantially mitigation of the DOIC problem when compared to SS-GANS and GANs without the SOMs. Furthermore, the SS-GANS performed much better than GANS on the DOIC problem, even without the SOMs.
\end{abstract}


\section{Introduction}

First introduced by Goodfellow et al. \cite{goodfellow2014generative}, generative adversarial networks are trained to generate fake but realistic data using unsupervised learning. Following that, Salimans et al. \cite{salimans2016improved} demonstrated how GANs could be used to perform semi-supervised learning.

Consider equations \ref{GAN_semi}-\ref{GAN_semi_unsuper} from Salimans et al. \cite{salimans2016improved}. These equations define the objective function for the generator and discriminator. 

\begin{equation}\label{GAN_semi}
J(\theta_D, \theta_G) = J(\theta_D)_{supervised} + J(\theta_D, \theta_G)_{unsupervised}
\end{equation}

\begin{equation}\label{GAN_semi_super}
\begin{split}
\max_{\theta_D \in \Theta_D} J(\theta_D)_{supervised} = \\
-\mathbb{E}_{\bm{x},\bm{y}\sim p_{x,y}(x,y)}[log p_{model}(\bm{y}|\bm{x}, \bm{y}<K+1)]
\end{split}
\end{equation}

\begin{equation}\label{GAN_semi_unsuper}
\begin{split}
\min_{\theta_G \in \Theta_G} \max_{\theta_D \in \Theta_D} J(\theta_D, \theta_G)_{unsupervised} = \\
-\{\mathbb{E}_{\bm{x} \sim p_{x}(x)}log[1-p_{model}(\bm{y}=K+1|\bm{x})] + \\
\mathbb{E}_{\bm{z} \sim p_{z}(z)}log[p_{model}(\bm{y}=K+1|G(\bm{z}))]\}
\end{split}
\end{equation}

This objective function consists of terms, shown in equation \ref{GAN_semi_super} for the labeled data, and in equation \ref{GAN_semi_unsuper} for unlabeled data. As is typical with GANs, $\Theta_D$ and $\Theta_G$ denote all possible parameter values of the discriminator and generator.  Let $p_x$ and $p_z$ denote the Probability Density Functions (PDFs) of the training samples and the generator noise distribution. Let $p_{model} = p(\bfy | \bfx_0; \theta_D)$ denote the conditional probability of assigning a label $\bfy$ given an input $\bfx_0$ and model parameters $\theta_D$. Equation \ref{GAN_semi_super} maximizes the expected value of the log probability of correctly classifying the labeled, inlier training samples.

Equation \ref{GAN_semi_unsuper} has two components, the first specifies that the real unlabeled data should not be classified as generated, and the second is the standard GAN adversarial term. This expression distinguishes real samples with unknown labels from generated data. The intention of this approach is that the unlabeled data can help $G$ to better learn the data distribution, which in turn will allow $D$ to perform better classification.

There are some other components of the work from \cite{salimans2016improved} which make it effective for semi-supervised classification. One is the authors' introduction of virtual batch normalization. More importantly is their usage of feature matching as the objective for the generator, shown in equation \ref{GAN_feature_matching}. 

\begin{equation}\label{GAN_feature_matching}
||\mathbb{E}_{x \sim p_{data}} f(x) - \mathbb{E}_{z \sim p_{z}(z)} f(G(z))||_{2}^{2}
\end{equation}

$f(x)$ in equation \ref{GAN_feature_matching} refers to the outputs from an intermediate layer of the discriminator. Instead of requiring the generator to produce samples that maximize the output of the discriminator, this objective only requires the generator to produce samples that have the correct statistics according to the internals of the discriminator. The authors argue for this method by pointing out that the generator should receive useful information by directly learning features the discriminator has found to be discriminative.

The work in \cite{salimans2016improved} makes an assumption that all unlabeled data is from the classes of interest. The term $-\{\mathbb{E}_{\bm{x} \sim p_{x}(x)}log[1-p_{model}(\bm{y}=K+1|\bm{x})]$ in equation \ref{GAN_semi_unsuper} implies that all data $\bm{x}$ given to the system is from one of the $K$ classes that $p_{model}$ is being trained to classify. We feel that this assumption is not valid in general for semi-supervised applications.

The goal of semi-supervised learning is to make use of a large, unlabeled data set. In practice, that unlabeled data may contain outliers. In order to make use of unlabeled data, and to have a robust classifier, we would like a competency aware classifier that can identify outlier points both in training and testing. We use the term outlier to encompass all samples that are not from the classes of interest, i.e. samples $\bfv \not\sim p_x$.

\section{Method}

\subsection{SOM Distance}

Self-organizing maps (SOMs) represent a feature space using a set of exemplar samples. Distance to these samples encodes a point's position in the space. For details of the SOM training algorithm, see Kohonen \cite{kohonen1990self}.

A number of previous works have used SOMs for dealing with outliers. Chiang et al. used SOMs to mitigate DIOC problems for handwritten word recognition \cite{chiang1997hybrid} \cite{chiang1997recognition}. Frigui et al. used them for the same purpose in a landmine detection algorithm that was deployed in Afghanistan \cite{frigui2006detection} \cite{frigui2008detection}. Munoz and Muruz{\'a}bal used SOMs for outlier detection in distorted characters and in milk container data \cite{munoz1998self}. Li et al. used them for outlier detection in biomedical and credit card applications \cite{li2020multi}. Shahreza used a method combining a SOM with particle swarm optimization for anomaly detection applied to forest fire detection \cite{shahreza2011anomaly}.

The algorithm presented here is unique; the SOM has never been combined with any type of GAN, supervised or unsupervised. Furthermore, in contrast to other published approaches, the SOM and the input feature vectors are both provided to the network. The combination of both sets of feature vectors proves to be necessary to handling the DIOC problem.

\begin{figure}[ht]
    \centering
    \includegraphics[width=0.47\textwidth]{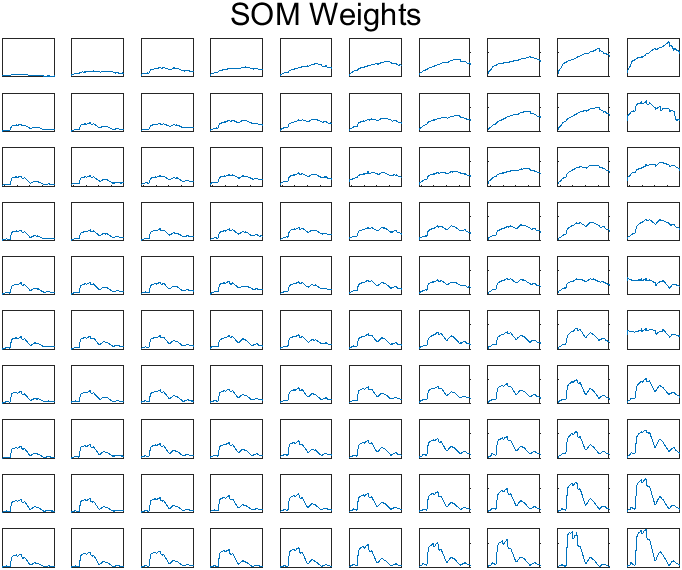}
    \caption{SOM weights trained with data from the Santa Barbara data described in Section \ref{sec:santa_barbara}}
    \label{fig:santabarbara_SOM}
\end{figure}

\begin{figure*}
     \centering
     \begin{subfigure}[b]{0.3\textwidth}
         \centering
         \includegraphics[width=\textwidth]{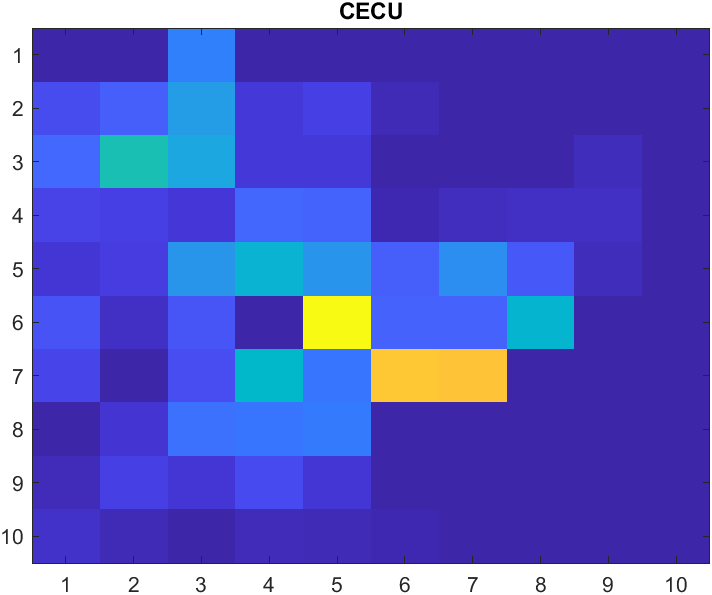}
         \caption{Ceanothus cuneatus}
         \label{fig:cecu_resp}
     \end{subfigure}
     \hfill
     \begin{subfigure}[b]{0.3\textwidth}
         \centering
         \includegraphics[width=\textwidth]{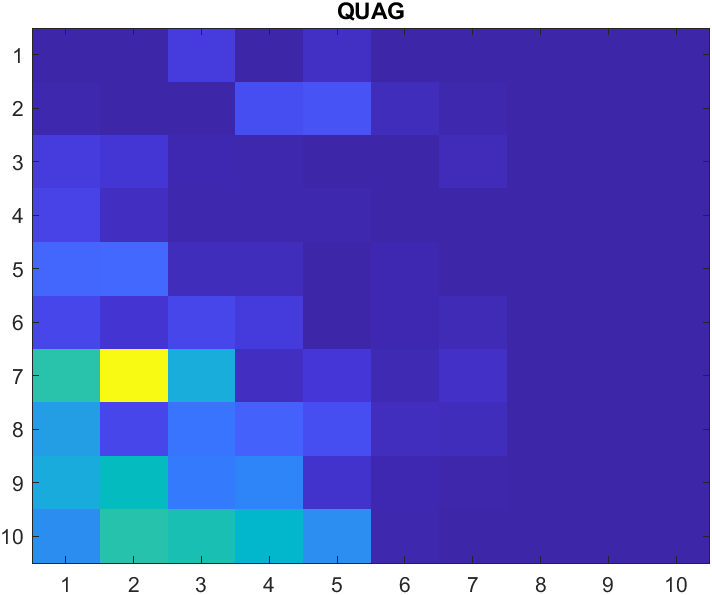}
         \caption{Quercus agrifolia}
         \label{fig:quag_resp}
     \end{subfigure}
     \hfill
     \begin{subfigure}[b]{0.3\textwidth}
         \centering
         \includegraphics[width=\textwidth]{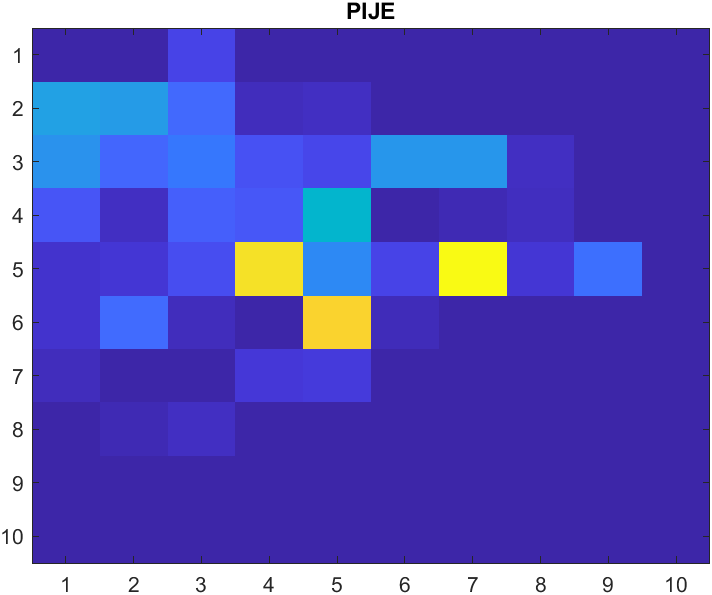}
         \caption{Pinus jeffreyi}
         \label{fig:pije_resp}
     \end{subfigure}
     \\
     \begin{subfigure}[b]{0.3\textwidth}
         \centering
         \includegraphics[width=\textwidth]{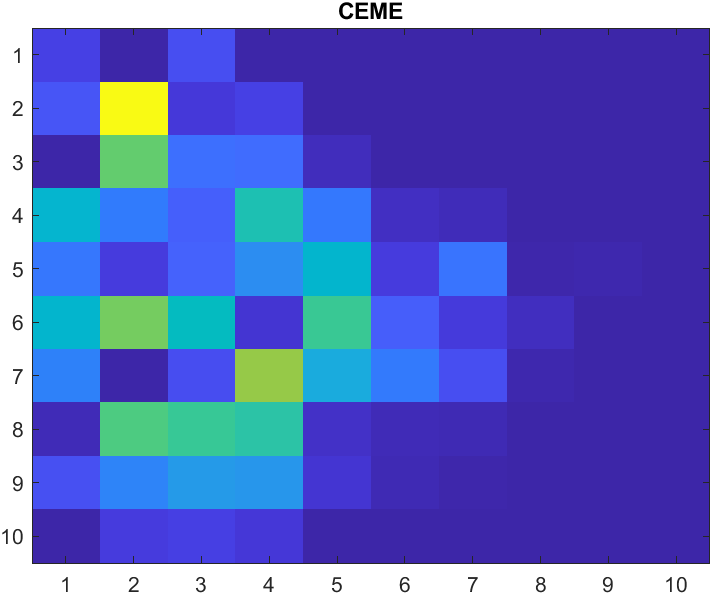}
         \caption{Ceanothus megacarpus}
         \label{fig:ceme_resp}
     \end{subfigure}
     \hfill
     \begin{subfigure}[b]{0.3\textwidth}
         \centering
         \includegraphics[width=\textwidth]{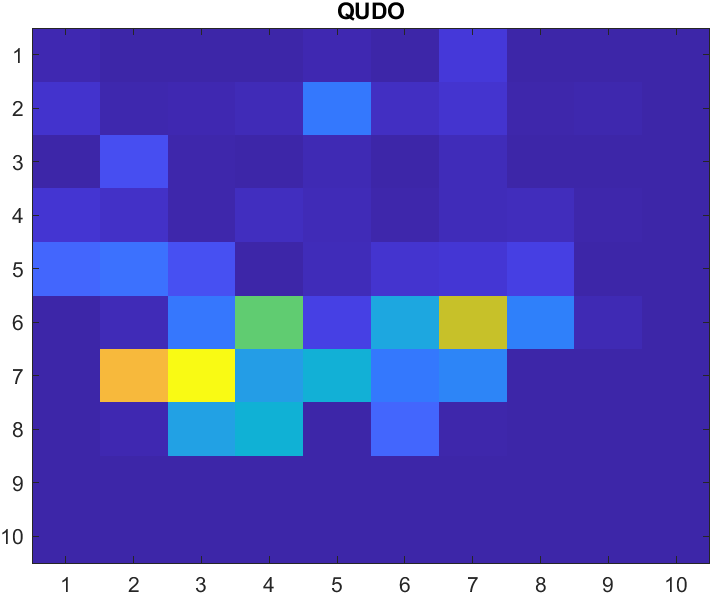}
         \caption{Quercus douglasii}
         \label{fig:qudo_resp}
     \end{subfigure}
     \hfill
     \begin{subfigure}[b]{0.3\textwidth}
         \centering
         \includegraphics[width=\textwidth]{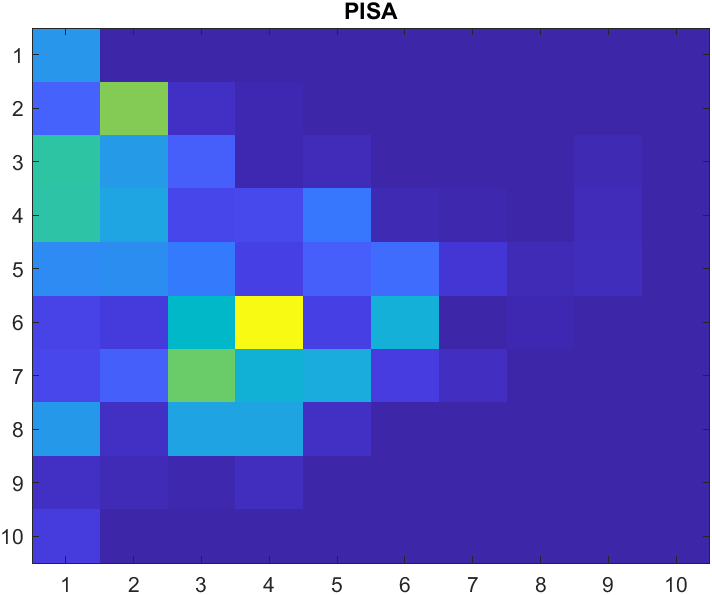}
         \caption{Pinus sabiniana}
         \label{fig:pisa_resp}
     \end{subfigure}
        \caption{Histograms showing several materials' response to the SOM in Figure \ref{fig:santabarbara_SOM}. These histograms show the frequency each node is the closest node to the samples from that class. Note the similarity in response of the map to the classes which are subspecies of the same species.}
        \label{fig:ambiguous_resps}
\end{figure*}

Figure \ref{fig:santabarbara_SOM} shows an example SOM. Note that the map attempts to represent the variability of the data, with shadows in the top left, bright sand in the top right, and various types of vegetation filling out the rest of the map. The map represents the ambiguity that exists between several of the classes in the data. Consider figure \ref{fig:ambiguous_resps}, which shows histograms of the closest node to the samples of several materials. The species which share a genus also share several nodes in the map, indicating the similarity of those materials in the SOM's representation.

SOM features are computed by taking an incoming sample and calculating its distance to each node in the SOM. This distance is generally taken to be euclidean distance. However, depending on the data source, euclidean distance may not be the most discriminative space.

In particular, in the case of spectra, euclidean distance has problems. The same material under different illumination conditions will be seperated by a large distance which is undesirable. Also, euclidean distance weighs differences in each feature the same. However, different wavelengths have vastly different ranges. A small change in the near-infrared range is likely not significant, but the same change in the visible range is likely to be relevant.

To address these issues, we use a combination of mahalanobis distance and spectral angle. Mahalanobis distance is shown in equation \ref{eq:mahal_dist}. The matrix $S$ refers to the covariance matrix formed from a set of observations. We compute a unique covariance $S$ for each node in the SOM based on the samples from the training set which are closest to those nodes in terms of euclidean distance. The use of mahalanobis distance accounts for the differences in variance between bands.

The formula for spectral angle is shown in equation \ref{eq:spec_angle}. Spectral angle was introduced by Kruse et al. \cite{kruse1993spectral}. It measures a distance between spectra which is illumination invariant. This allows materials like sand, which can have large illumination variance, to be close. We use a combination of mahalanobis distance and spectral angle as shown in equation \ref{eq:SOM_dist}.

\begin{equation}\label{eq:mahal_dist}
D_{mahalanobis}(x, y) = \sqrt{(x - y)^{T}S^{-1}(x - y)}
\end{equation}

\begin{equation}\label{eq:spec_angle}
D_{spectral angle}(x, y) = \arccos{\frac{x \cdot y}{||x|| \cdot ||y||}}
\end{equation}

\begin{equation}\label{eq:SOM_dist}
D^{*} = D_{mahalanobis} + 40*D_{spectral angle}
\end{equation}

\begin{figure}[ht]
    \centering
    \includegraphics[width=0.47\textwidth]{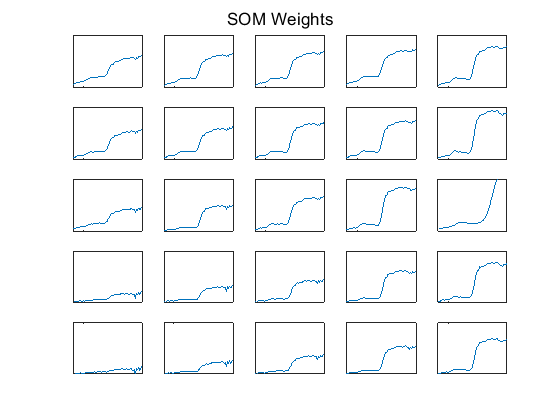}
    \caption{5x5 SOM weights trained with spectra from the tree and mostly grass classes.}
    \label{fig:SOM5x5}
\end{figure}

\begin{figure}[ht]
    \centering
    \includegraphics[width=0.47\textwidth]{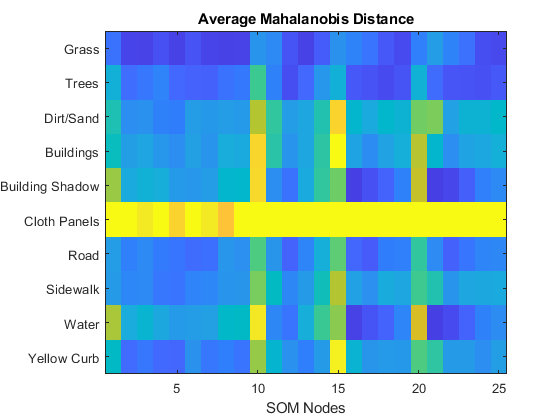}
    \caption{Average Mahalanobis distance between materials from the MUUFL dataset and SOM nodes. The trees and mostly grass materials are the inlier materials.}
    \label{fig:avgMahalDists}
\end{figure}

\begin{figure}[ht]
    \centering
    \includegraphics[width=0.47\textwidth]{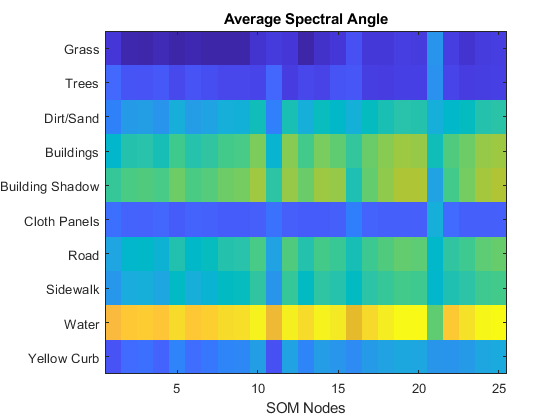}
    \caption{Average spectral angle between materials from the MUUFL dataset and SOM nodes. The trees and mostly grass materials are the inlier materials.}
    \label{fig:avgSpecAngle}
\end{figure}

\begin{figure}[ht]
    \centering
    \includegraphics[width=0.47\textwidth]{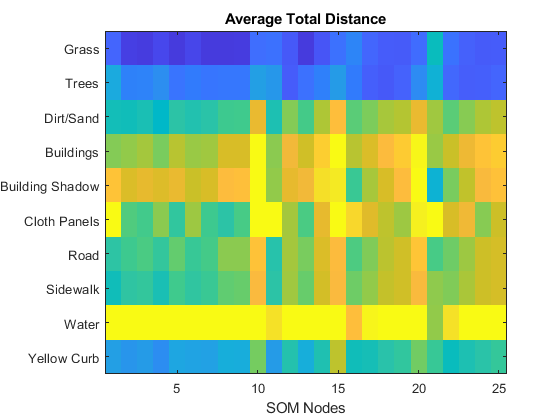}
    \caption{Average $D^{*}$ between materials from the MUUFL dataset and SOM nodes. The trees and mostly grass materials are the inlier materials.}
    \label{fig:avgTotalDists}
\end{figure}

Figures \ref{fig:SOM5x5}-\ref{fig:avgTotalDists} show an example of the distances. Figure \ref{fig:SOM5x5} shows a 5x5 SOM trained with spectra from the trees and mostly grass classes. These nodes represent the variability present in both materials. The top right of the map represents very bright vegetation. As you move towards the bottom left, you encounter darker vegetation and eventually shadow.

Figure \ref{fig:avgMahalDists} shows the average mahalanobis distance calculated between various materials and the nodes in the the map from figure \ref{fig:SOM5x5}. Grass is closer than all other materials to the bright vegetation nodes, while the the trees are close to the darker vegetation. Note that when looking at mahalanobis distance alone, there is some confusion between inlier and outlier materials. The building shadow and water materials are closer to the darkest nodes in the map than the trees or mostly grass materials are.

Figure \ref{fig:avgSpecAngle} shows the average spectral angle between materials and the nodes in the SOM. The inlier materials (mostly grass and trees) tend to be closer than the outlier materials. Again, there is some confusion with other materials, particularly the cloth panels. However, the materials which are confused in spectral angle are different than those that are confused in mahalanobis distance. The building shadow and water materials, which are close to some of the nodes in the SOM in terms of mahalanobis distance, are relatively far from those nodes in terms of spectral angle. The resulting $D^{*}$ from equation \ref{eq:SOM_dist} is shown in figure \ref{fig:avgTotalDists}.

\subsection{SOM Outlier Detection} \label{sec:som_outlier}

We would like to use the SOM to detect outliers. Outliers should have large distances to all nodes in the map as they are unlike the legitimate data. To facilitate detection, we would like all outliers to have similar features after being presented to the map. We do this by converting the distances obtained from the map into membership values for each node.

SOM membership values were introduced by Chiang and Gader \cite{chiang1997recognition} We follow a similar approach in this work. Each node in the map is paired with a sigmoid unit of the form in equation \ref{eq:SOM_sigmoid}. $d$ refers to the distance obtained by comparing an incoming sample with the node, and $\alpha$ and $\beta$ are learnable parameters.

\begin{equation}\label{eq:SOM_sigmoid}
f(d, \alpha, \beta) = \frac{1}{1 + e^{\alpha(d - \beta)}}
\end{equation}

The goal of each sigmoid is to map inlier samples close to that node to large membership values. All other inlier samples, as well as all outlier samples, should be mapped to small membership values. The parameters $\alpha$ and $\beta$ are trained towards this goal.

For each $i^{th}$ sample in the training set, we define a target membership value $t$ for each $j^th$ node in the map. To do this, we take each sample in the training set and find it's best matching unit (BMU) in the map. The target values $t$ for that sample are then set with a value of 1 at the BMU, and decaying around it. The target values are shown in equation \ref{eq:SOM_targetvalues}.

\begin{equation}\label{eq:SOM_targetvalues}
    t_{ij}(x)= 
\begin{cases}
    1,              & \text{if } j \text{ is BMU}\\
    0.5,              & \text{elif } j \text{ is direct neighbor of BMU}\\
    0.25,              & \text{elif } \text{ j is in 9-neighborhood of BMU}\\
    0,              & \text{otherwise}
\end{cases}
\end{equation}

Consider $O_{ij}$ to be the output of equation \ref{eq:SOM_sigmoid} at node $j$ for sample $i$. We penalize the outputs using mean squared error, as shown in equation \ref{eq:SOM_targetobj}. The parameters $\alpha$ and $\beta$ are updated to minimize this objective.

\begin{equation}\label{eq:SOM_targetobj}
    E = \sum_{i}{[\frac{1}{2}{\sum_{j}{(t_{ij} - O_{ij})^{2}}}]}
\end{equation}

\subsection{Architecture}

\begin{figure}[ht]
    \centering
    \includegraphics[width=0.47\textwidth]{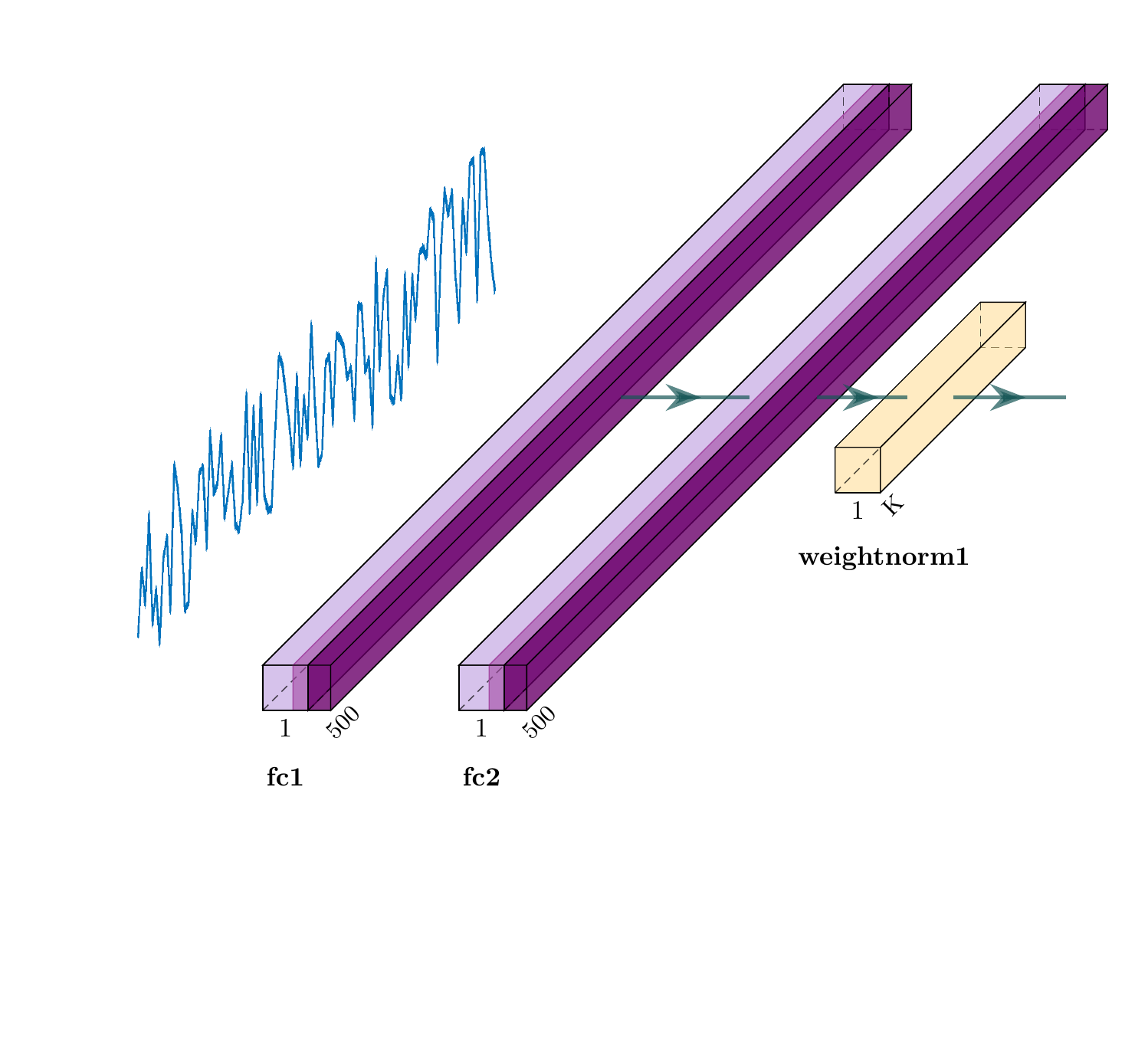}
    \caption{Architecture of the generator network.}
    \label{fig:gen_arch}
\end{figure}

The architecture used in this work follows the architecture used by Salimans et al. \cite{salimans2016improved}. Figure \ref{fig:gen_arch} shows the architecture of the generator network. This network takes uniform random noise as input and outputs a generated spectra. The first two layers are fully connected layers with batch normalization and leaky ReLU activation unites. The final layer is a weight normalized fully connected layer, as described in \cite{salimans2016weight}.

\begin{figure*}
    \centering
    \includegraphics[width=1\textwidth]{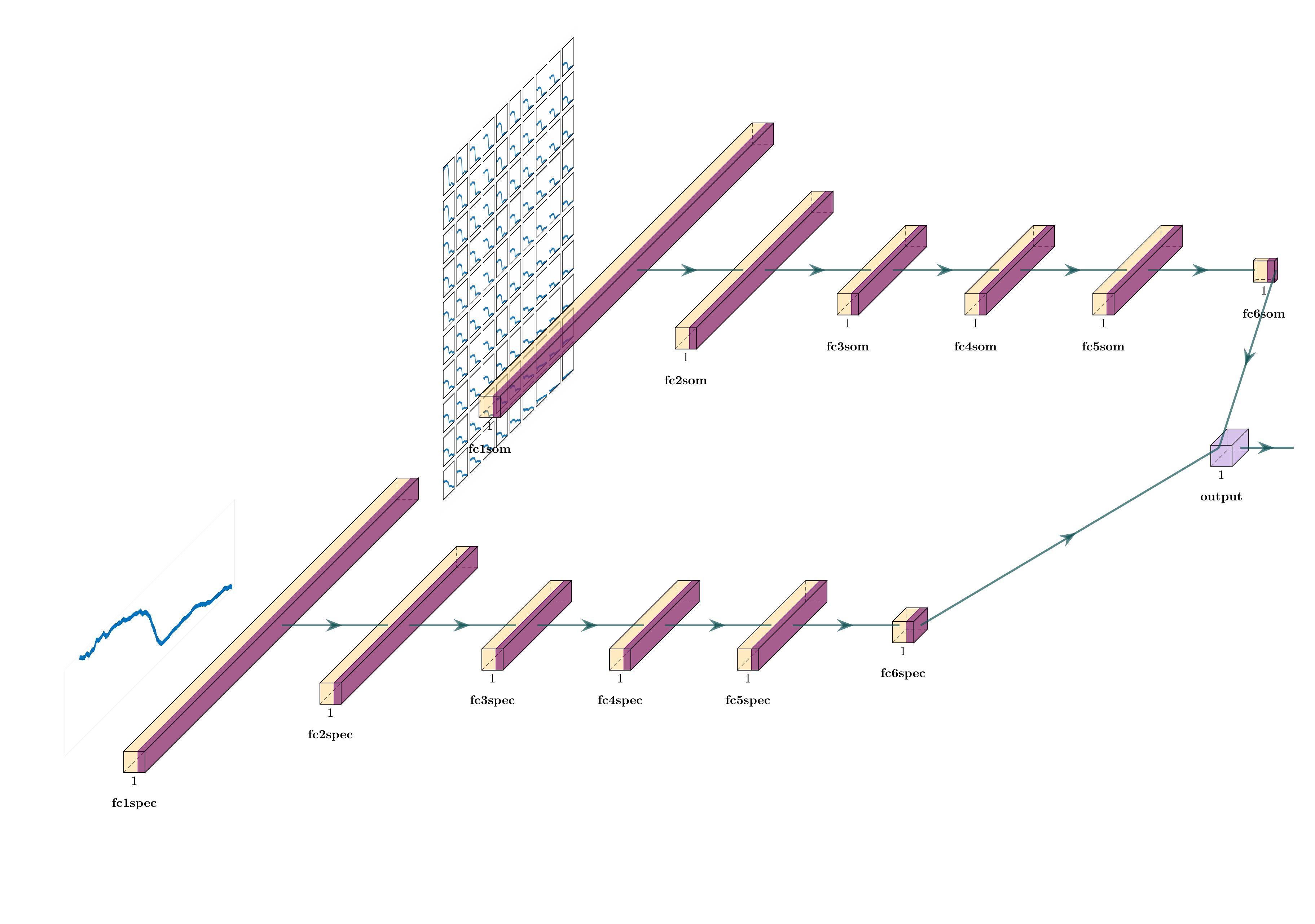}
    \caption{Architecture of the discriminator network.}
    \label{fig:disc_arch}
\end{figure*}

The discriminator architecture is shown in figure \ref{fig:disc_arch}. The discriminator takes both a spectra and a map of SOM features as input. The SOM features are computed according to the method outlined in section \ref{sec:som_outlier}. The two sets of features are passed through separate paths of the network. The layers are formed from weight normalized fully connected units with leaky ReLU activation functions.

The two different paths through the network are used for different purposes. SOM features are specifically designed for outlier detection, as described in section \ref{sec:som_outlier}, so that section of the network is trained to perform outlier detection. The section of the network using the spectral features is trained to perform classification. The outputs of both sections of the network are concatenated in the final layer.

\section{Data sets}

\subsection{MUUFL Gulfport}

\begin{figure*}
    \centering
    \includegraphics[width=1\textwidth]{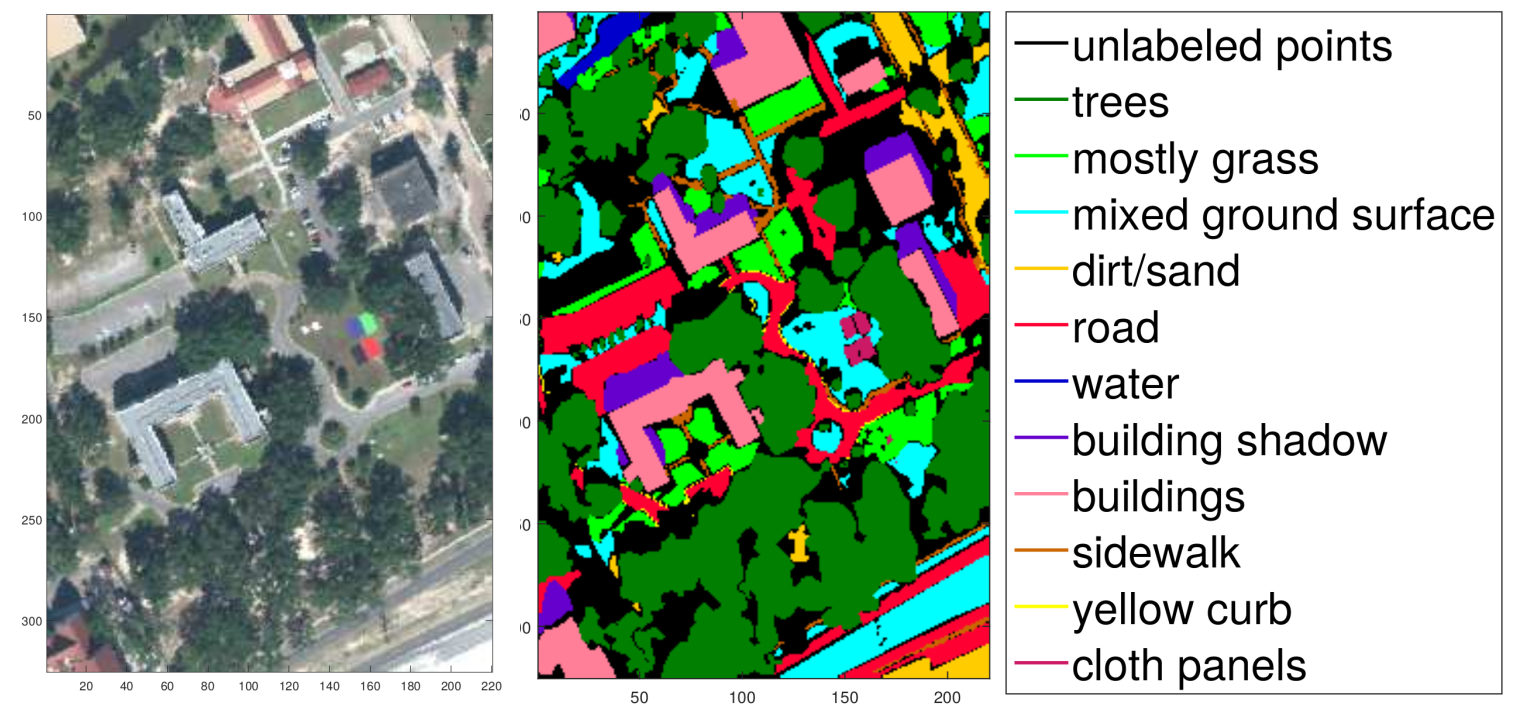}
    \caption{MUUFL scene and ground truth from Gader et al. \cite{gader2013muufl}}
    \label{fig:muufl_labeling}
\end{figure*}

The MUUFL Gulfport dataset contains hyperspectral data collected over Gulfport, MS in 2013. \cite{gader2013muufl} The data consists of the visible and near-infrared range from 368-1043 nm. It contains a variety of materials vegetation and manmade. Figure \ref{fig:muufl_labeling} shows the labeling performed by Gader et al. \cite{gader2013muufl}.

The goal of this study was to correctly classify the grass and tree classes, while considering the following classes as outliers: road materials, water, building shadow, yellow curb, cloth panels, dirt and sand, buildings, and sidewalk.

\subsection{Santa Barbara AVIRIS} \label{sec:santa_barbara}

Details of the Santa Barbara AVIRIS dataset are provided in Meerdink et al. \cite{meerdink2019classifying} We summarize relevant details of that discussion here. The dataset contains images collected during the HyspIRI Airborne Preparatory Campaign. These images were collected using the AVIRIS sensor, which operates between 360 and 2500 nm, as discussed in Green et al. \cite{green1998imaging} Data was collected by flying the sensor on the NASA ER-2 aircraft over six flightboxes; the flightbox focused on here is the Santa Barbara flightbox.

\begin{figure}[ht]
    \centering
    \includegraphics[width=0.47\textwidth]{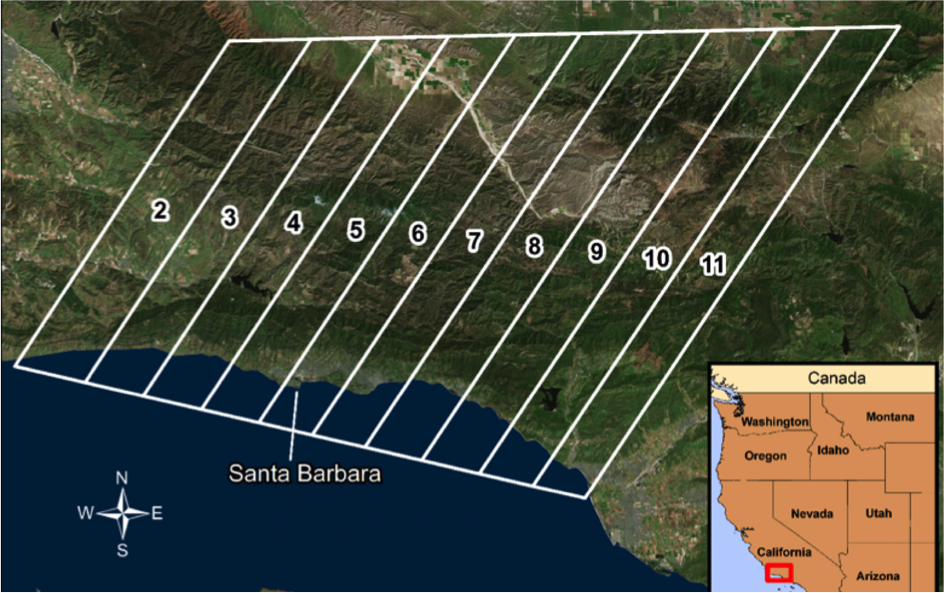}
    \caption{AVIRIS HyspIRI Flight Lines (image from \cite{meerdink2019classifying}).}
    \label{fig:HyspIRIFlightLines}
\end{figure}

The Santa Barbara flightbox is pictured in figure \ref{fig:HyspIRIFlightLines}. These flightlines cover a range environments, from coastal Santa Barbara to the Los Padres National Forest. This data consists of a time series which was collected three times per year in April, June, and November or August during 2013, 2014, and 2015. We conduct experiments here with the data from Spring 2013.

Reference data was collected for this dataset by Meerdink et al. \cite{meerdink2019classifying} This reference data was collected using a rule that patches must contain at least 75\% of a given species to be labeled as that species. The reference data was used to create polygons that can be compared with the AVIRIS remote sensing data. The reference data contains 30 classes, which are URBAN, SOIL, ROCK, and 27 different tree species. In our experiments, we include all of these materials for classification with the exception of the URBAN class. For outliers, we manually label several other materials that exist in the scene. These materials are: rooftops, sand, water and clouds.

\section{Experimental Results}

\begin{figure}[ht]
    \centering
    \includegraphics[width=0.47\textwidth]{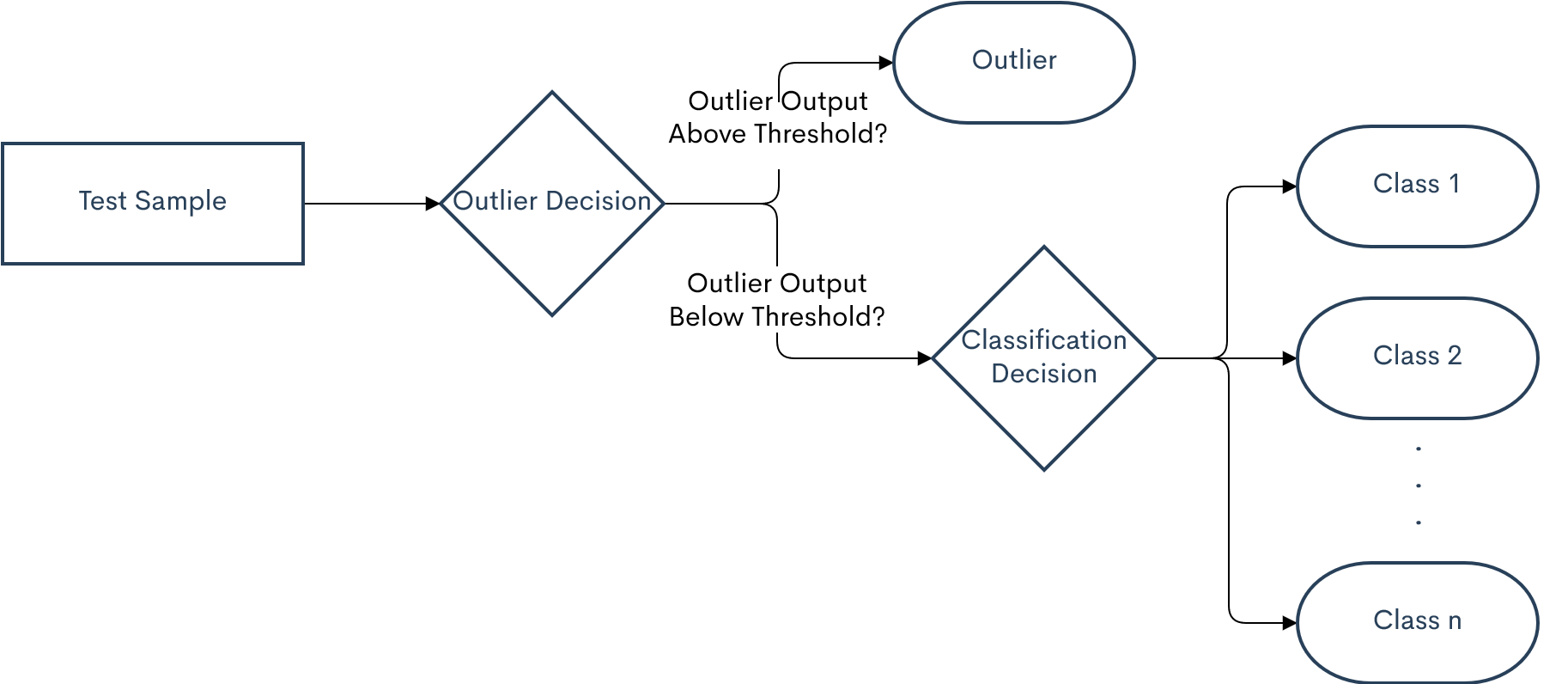}
    \caption{Flowchart of decisions for test samples.}
    \label{fig:outlier_flowchart}
\end{figure}

In this work, we consider test samples that may be outliers or inliers, as well as classification performance within the inlier samples. Based on this, we classify test samples with a two-step process shown in figure \ref{fig:outlier_flowchart}. We first use the model to decide if a sample is outlier or inlier, and then if that sample is classifed as inlier, then we further classify it into one of several classes.

Experiments were done on the MUUFL data set. Models were trained to identify two classes: trees and mostly grass. Outliers were considered to be samples from all other classes: road materials, water, building shadow, yellow curb, cloth panels, dirt and sand, buildings, and sidewalk.

The labeled training set was composed of 10 samples from each of the trees and mostly grass classes. Additionally, 10 training outlier samples were provided. These samples corresponded to samples taken from the cloth panels originally placed into the scene for calibration. The unlabeled training set was composed of 500 samples from each of those classes as well as 3500 samples drawn from the group of outlier materials described above. This sampling was repeated 20 different times.

Four different model types were trained to understand the effects of supervised vs semi-supervised objectives as well as spectral features vs spectral and SOM features. We train two models with only the supervised component from equation \ref{GAN_semi}, one with only spectral features and the other with spectral and SOM features. The other two models use the full objective from equation \ref{GAN_semi}, and again one uses only spectral features and the other uses the spectra and the SOM output.

\begin{figure*}
     \centering
     \begin{subfigure}[b]{0.3\textwidth}
         \centering
         \includegraphics[width=\textwidth]{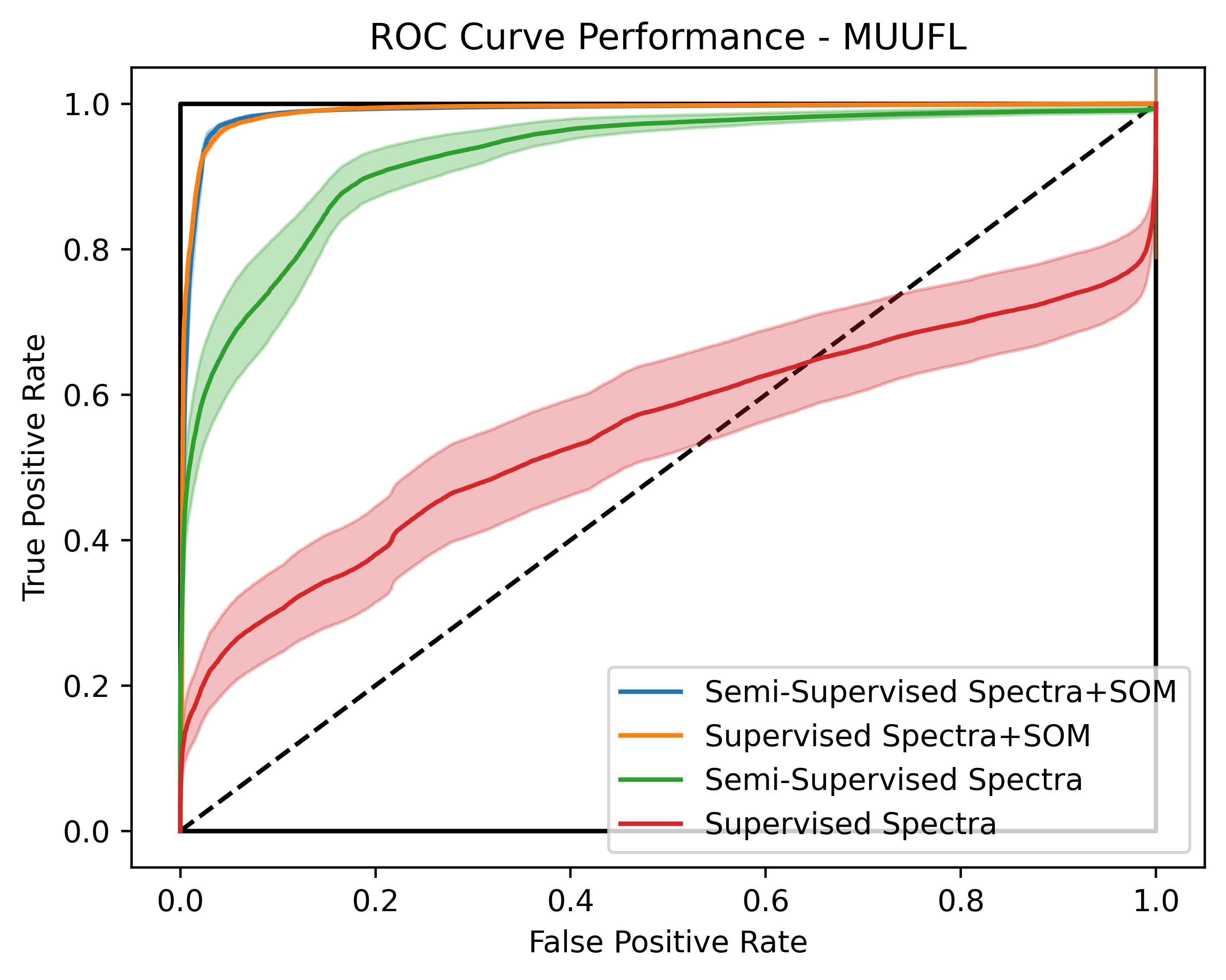}
         \caption{MUUFL ROC}
         \label{fig:all_models_roc_muufl}
     \end{subfigure}
     \hfill
     \begin{subfigure}[b]{0.3\textwidth}
         \centering
         \includegraphics[width=\textwidth]{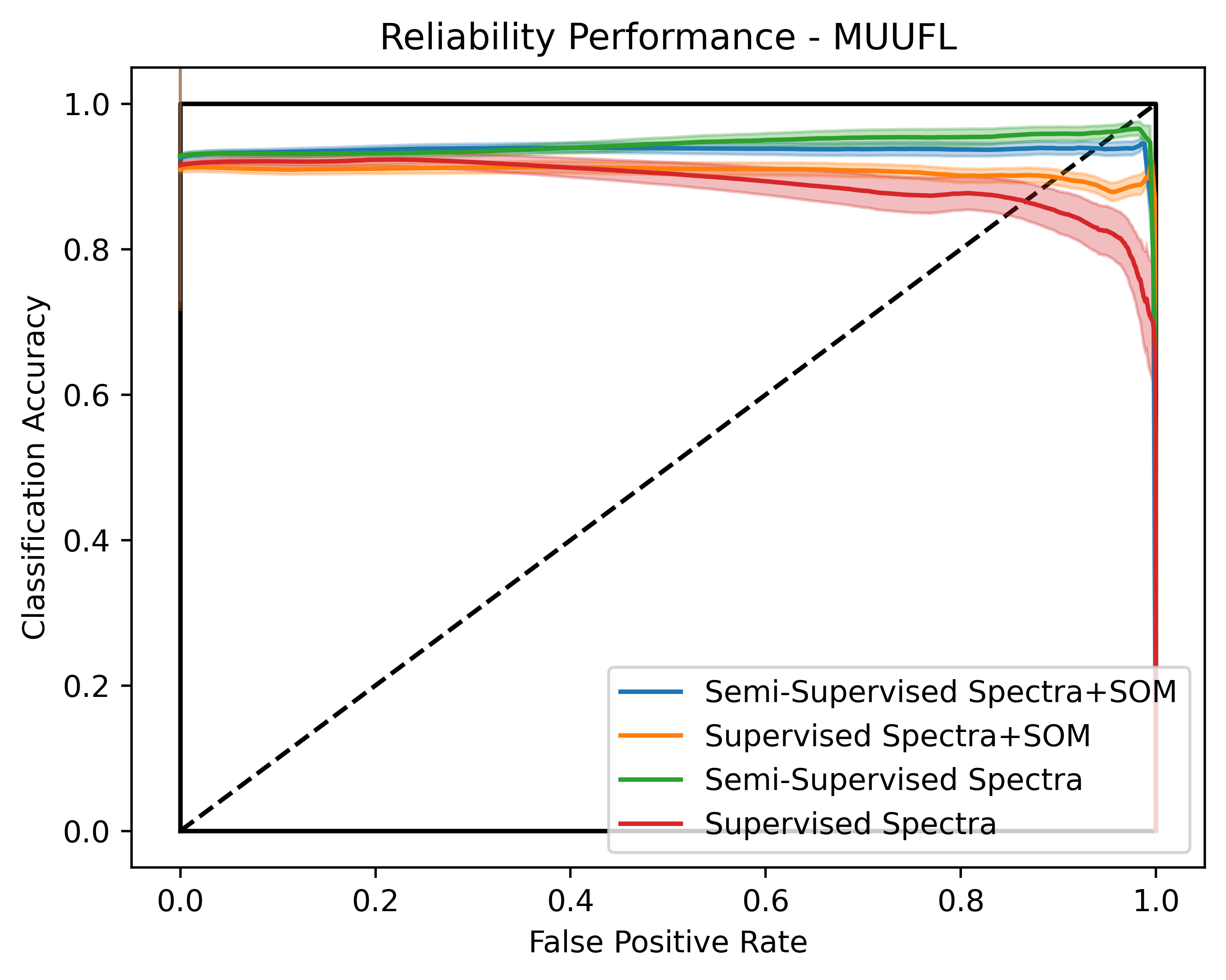}
         \caption{MUUFL Reliability}
         \label{fig:all_models_rel_muufl}
     \end{subfigure}
     \hfill
     \begin{subfigure}[b]{0.3\textwidth}
         \centering
         \includegraphics[width=\textwidth]{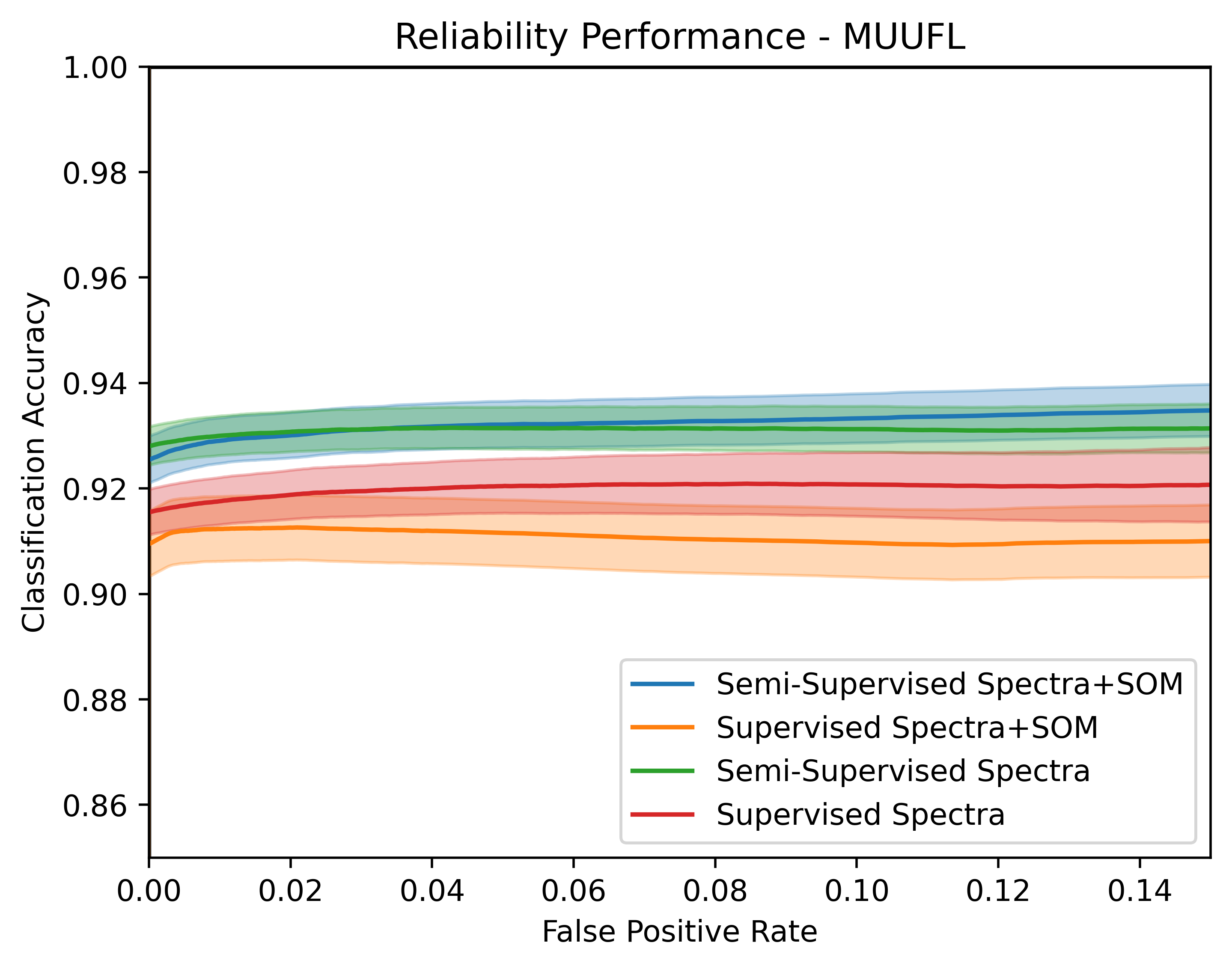}
         \caption{MUUFL Reliability Zoomed-In}
         \label{fig:all_models_rel_zoomed_muufl}
     \end{subfigure}
     \\
     \begin{subfigure}[b]{0.3\textwidth}
         \centering
         \includegraphics[width=\textwidth]{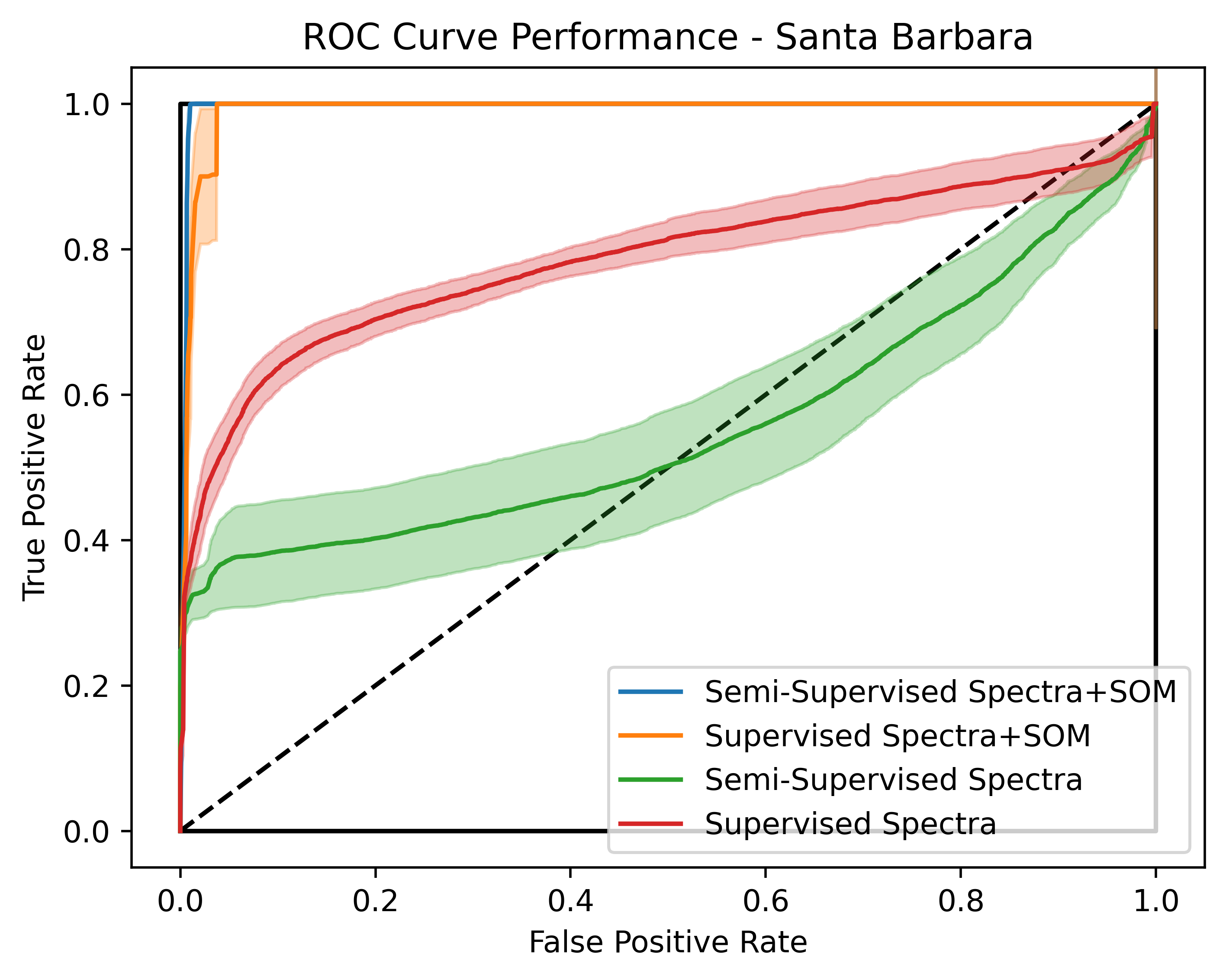}
         \caption{Santa Barbara ROC}
         \label{fig:all_models_SantaBarbara_roc}
     \end{subfigure}
     \hfill
     \begin{subfigure}[b]{0.3\textwidth}
         \centering
         \includegraphics[width=\textwidth]{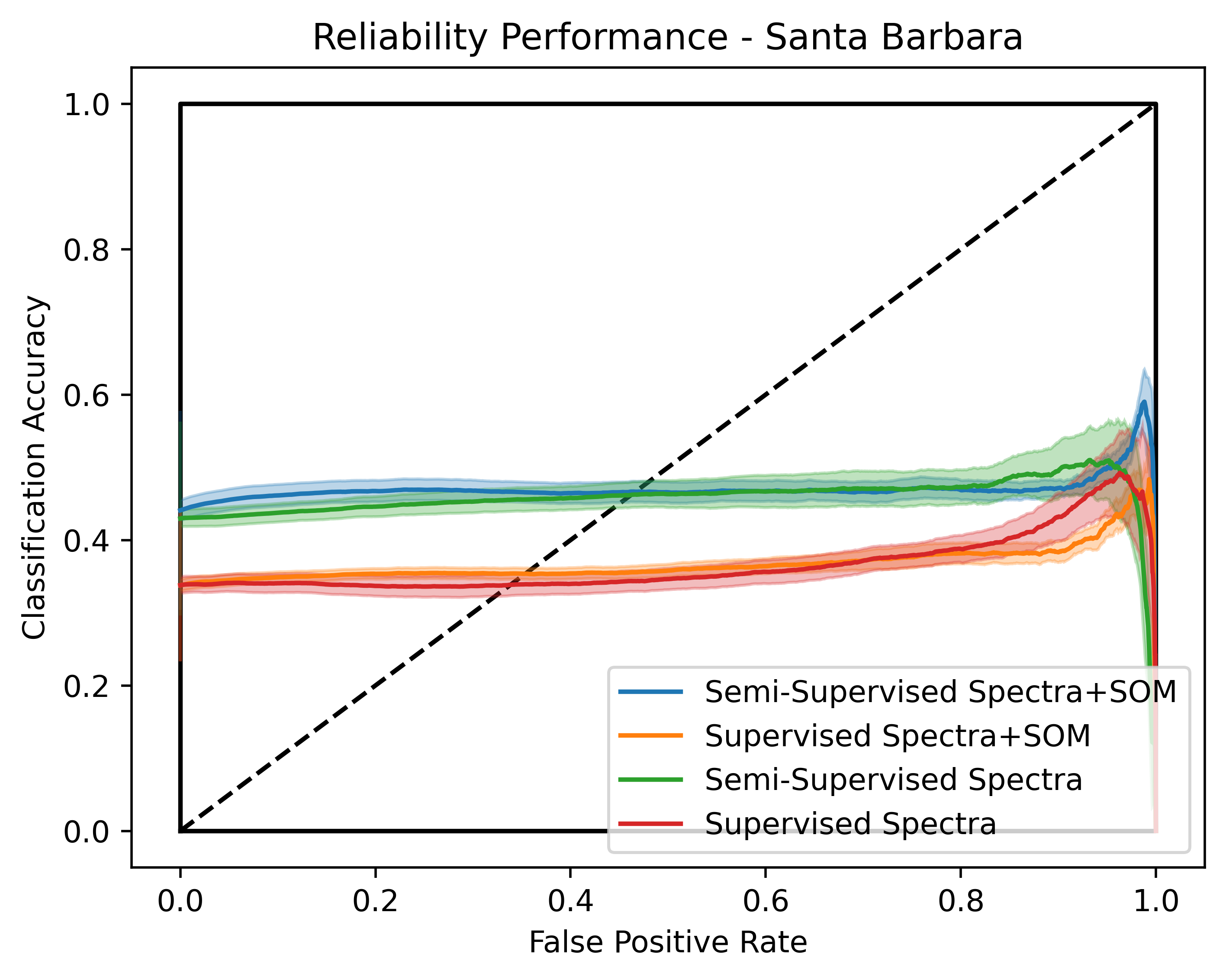}
         \caption{Santa Barbara Reliability}
         \label{fig:all_models_SantaBarbara_rel}
     \end{subfigure}
     \hfill
     \begin{subfigure}[b]{0.3\textwidth}
         \centering
         \includegraphics[width=\textwidth]{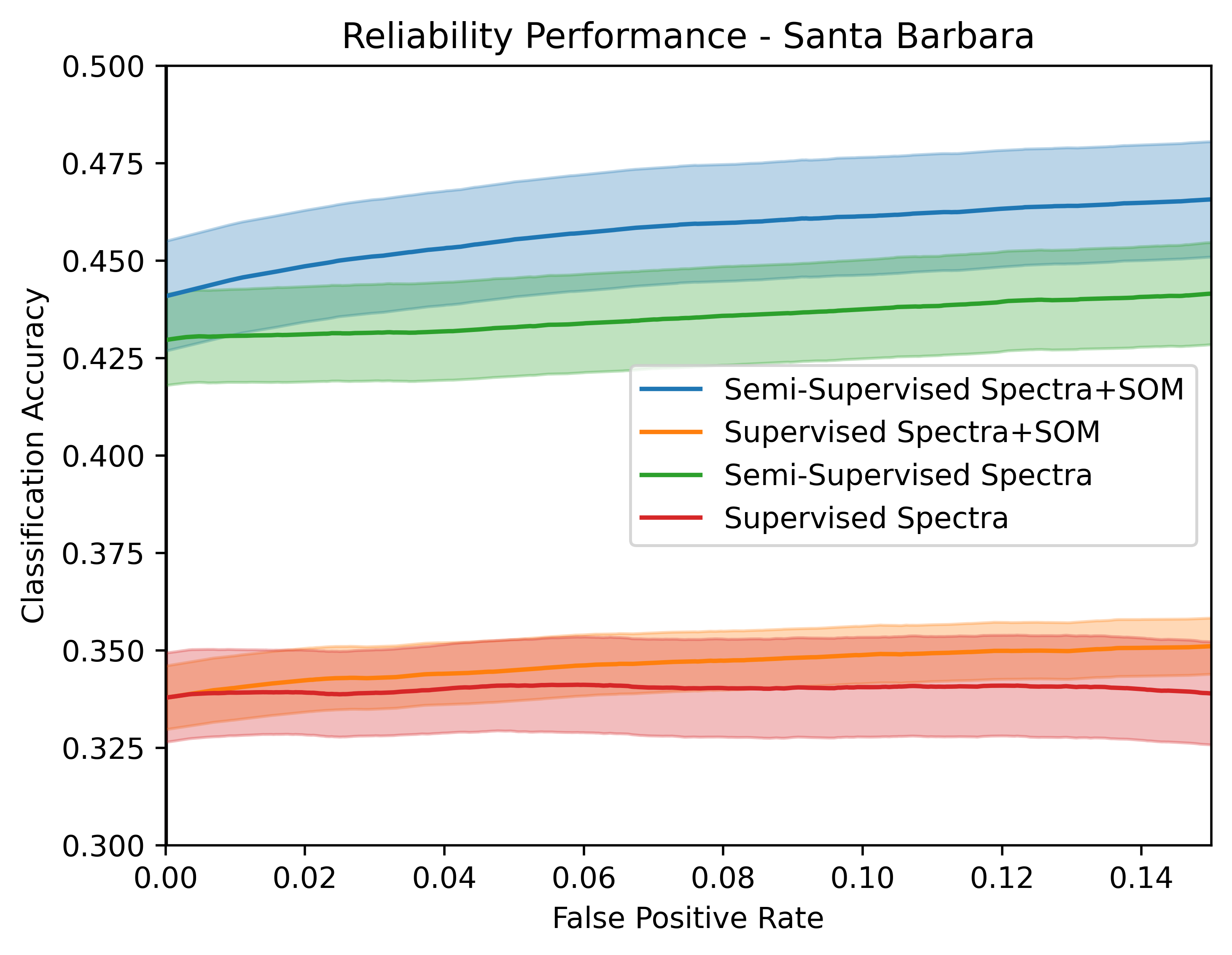}
         \caption{Santa Barbara Reliability Zoomed-In}
         \label{fig:all_models_SantaBarbara_rel_zoomed}
     \end{subfigure}
        \caption{Model performance on both the MUUFL Gulfport and Santa Barbara datasets. Results are shown with ROC curves and reliability scores. Mean values are shown in bold lines and the shaded areas indicate a 95\% confidence interval.}
        \label{fig:all_models}
\end{figure*}

Figures \ref{fig:all_models_roc_muufl} and \ref{fig:all_models_rel_muufl} show the results of these four models in terms of outlier detection ability and reliability across a range of detection thresholds. Figure \ref{fig:all_models_rel_zoomed_muufl} shows the reliability results zoomed in on the performance at low false alarm rates.

These figures show the outlier detection ability of each model in ROC curves, with the mean ROC curve shown in the bold line, and a 95\% confidence interval around that mean in the shaded area. Also shown is the classification performance shown in the dashed curves, again with a 95\% confidence interval around them. Both the ROC curves and classification performance are shown at various false alarm rates, implying that both metrics depend on the outlier threshold, as shown in figure \ref{fig:outlier_flowchart}.

Equation \ref{eq:reliability} shows the way that classification accuracies are computed. $N{a}$ refers to the number of inliers available for classification. Those are the inlier samples which were below the outlier threshold as shown in figure \ref{fig:outlier_flowchart}. $N_{c}$ refers to the number of those inlier samples which were correctly classified after having been correctly identified as inliers.

\begin{equation}\label{eq:reliability}
    acc(x)= 
    \begin{dcases}
        \frac{N_{c}}{N_{a}},& \text{if } N_{a}\geq 1\\
        0,              & \text{otherwise}
    \end{dcases}
\end{equation}

The results demonstrate the value of both the semi-supervised training as well as the SOM features. The ROC curves show that including the SOM features has a large impact in terms of outlier detection. The models that only use spectral features have very poor outlier detection, while the models using both spectral and SOM features are nearly perfect on that task. In terms of the secondary classification problem, the semi-supervised models have a 1-2\% increase in terms of classification accuracy. The semi-supervised model using both spectral and SOM features achieves both of these results simultaneously. Table \ref{tab:muufl} summarizes these results.

\begin{figure*}
    \centering
    \includegraphics[width=1\textwidth]{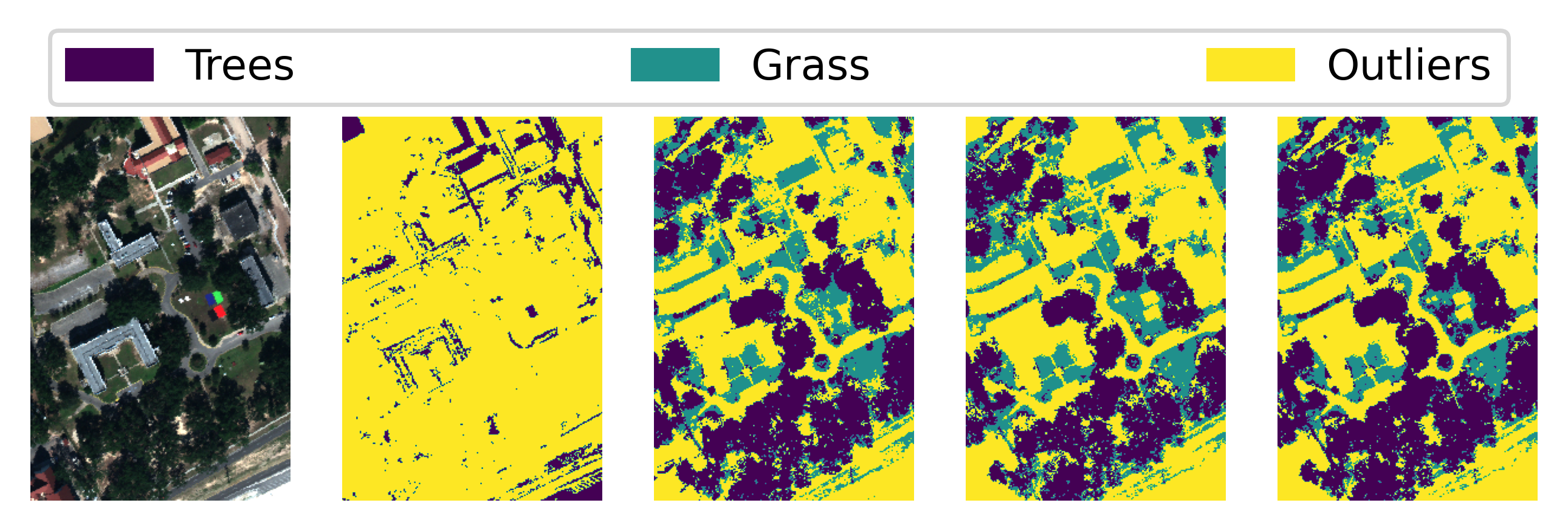}
    \caption{Full image outputs for various methods. All thresholds for outlier detection were selected to reject 90\% of outliers. Pictured from left to right: (1) the RGB of the MUUFL Gulfport dataset, (2) Supervised with spectral features only, (3) Semi-supervised GAN with spectral features only, (4) Supervised with spectral and SOM features, (5) Semi-supervised with spectral and SOM features}
    \label{fig:full_image_outputs}
\end{figure*}

Figure \ref{fig:full_image_outputs} shows the full image outputs for the different methods. In each case, the threshold for outlier detection was selected to reject 90\% of outliers. Pixel assignment was done by taking the most common classification across 10 models.

The supervised method with only spectral features in figure \ref{fig:full_image_outputs}-(2) lacks any meaningful outlier detection ability. The semi-supervised GAN with spectral features only in figure \ref{fig:full_image_outputs}-(3) still has a weakness in outlier detection. This can be seen by looking at it's predictions for the cloth panels in the scene, which it frequently mislabels as grass. Figures \ref{fig:full_image_outputs}-(4) and \ref{fig:full_image_outputs}-(5) which show the purely supervised method using the spectral and SOM features, and the semi-supervised method using spectral and SOM features, are both similar in terms of outlier performance. Their difference comes in terms of classification accuracy, which can be seen mainly in the clump of trees in the bottom of the images.

\begin{table}[h!]
\centering
\begin{adjustbox}{max width=\textwidth}
\begin{tabular}{ |c||c|c|  }
 \hline
 Model Type&Mean ROC AUC&\thead{Mean Top\\ Classification Rate}\\
 \hline
 \thead{Supervised\\ Spectra}&[0.491, 0.606]&[0.918, 0.928]\%\\
 \thead{Supervised\\ Spectra+SOM}&\textbf{[0.989, 0.990]}&[0.907, 0.919]\%\\
 \thead{Semi-Supervised\\ Spectra}&[0.908, 0.943]&\textbf{[0.929, 0.937]}\%\\
 \thead{Semi-Supervised\\ Spectra+SOM}&\textbf{[0.988, 0.990]}&\textbf{[0.928, 0.937]}\%\\
 \hline
\end{tabular}
\end{adjustbox}
\caption{Results for the MUUFL dataset. Results are given in terms of the mean area under the curve (AUC) of the ROC curve, as well as the mean of the top classification rates that the models attain between 0 and 0.05 false alarm rate. The range of values indicates a 95\% confidence interval.}
\label{tab:muufl}
\end{table}

Experiments were also done on the Santa Barbara data set. Models were trained to identify 29 of the classes from the reference dataset described in section \ref{sec:santa_barbara} with the only class excluded being the URBAN class. Outliers were considered to be samples from other materials in the scene: rooftops, sand, clouds, and water.

The labeled training set was composed of 10 samples from each of the 29 labeled classes. Additionally, 10 training outlier samples were provided, which were taken from the cloud spectra. The unlabeled training set was composed of 500 samples from each of those classes as well as 3500 samples drawn from the group of outlier materials described above. This sampling was repeated 10 different times.

The results on the Santa Barbara data are shown in figures \ref{fig:all_models_SantaBarbara_roc}-\ref{fig:all_models_SantaBarbara_rel_zoomed}. These results mirror the results on the MUUFL Gulfport data. Again, the models making use of the SOM features have significantly better outlier detection. In terms of classification performance, both semi-supervised models have roughly 10\% to 12\% higher classification accuracy. Table \ref{tab:santabarbara} summarizes these results.

Additionally, we include a comparison with the canonical discriminant analysis (CDA) method that was suggested by Meerdink et al \cite{meerdink2019classifying} for this problem. That method is only intended for classification, so we compare its performance to the models presented here at 0 false alarm rate. The CDA model results are shown in table \ref{tab:santabarbara} alongside the results from the GAN models. Both semi-supervised models we present outperform the CDA model in terms of classification accuracy, demonstrating the benefit of the semi-supervision.

\begin{table}[h!]
\centering
\begin{adjustbox}{max width=\textwidth}
\begin{tabular}{ |c||c|c|  }
 \hline
 Model Type&Mean ROC AUC&\thead{Mean Top\\ Classification Rate}\\
 \hline
 \thead{Supervised\\ Spectra}&[0.761, 0.808]&[0.333, 0.356]\%\\
 \thead{Supervised\\ Spectra+SOM}&[0.988, 0.994]&[0.337, 0.353]\%\\
 \thead{Semi-Supervised\\ Spectra}&[0.497, 0.617]&[0.423, 0.447]\%\\
 \thead{Semi-Supervised\\ Spectra+SOM}&\textbf{[0.995, 0.996]}&\textbf{[0.441, 0.470]\%}\\
 \thead{CDA\\ Spectra}&n/a&[0.387, 0.400]\%\\
 \hline
\end{tabular}
\end{adjustbox}
\caption{Results for the Santa Barbara dataset. Results are given in terms of the mean area under the curve (AUC) of the ROC curve, as well as the mean of the top classification rates that the models attain between 0 and 0.05 false alarm rate. The range of values indicates a 95\% confidence interval.}
\label{tab:santabarbara}
\end{table}

\section{Conclusion}

A method was presented to allow semi-supervised GANs to handle outliers present in the unlabeled dataset, as well as during test. This method takes advantage of the self-organizing map's ability to map outlier samples to uniform non-response values. The system was tested on hyperspectral data from the MUUFL Gulfport dataset as well as from the HyspIRI Santa Barbara dataset. In both datasets, the semi-supervised training provided boosts in classification accuracy. The combination of spectral and SOM features was demonstrated to not lose any classification performance vs pure spectral features, but gain significant performance in terms of robustness vs outliers.

\bibliographystyle{unsrt}
\bibliography{bib/references}

\end{document}